\pdfoutput=1

\documentclass[11pt]{article}
\usepackage{acl2016}
\usepackage{times}
\usepackage{latexsym}
\usepackage{xspace}
\usepackage{arydshln}
\usepackage{graphicx}
\usepackage{multirow}
\usepackage{amsmath}
\usepackage{relsize}
\usepackage{url}
\usepackage{caption}
\usepackage{subcaption}
\usepackage{adjustbox}
\usepackage{amsfonts,amssymb}

\newcommand{\method}[1]{\texttt{#1}\xspace}
\newcommand{\doctovec}{\method{doc2vec}}
\newcommand{\dbow}{\method{dbow}}
\newcommand{\dmpv}{\method{dmpv}}
\newcommand{\ngram}{\method{ngram}}
\newcommand{\wordtovec}{\method{word2vec}}
\newcommand{\skipthought}{\method{skip-thought}}
\newcommand{\pp}{\method{pp}}
\newcommand{\glove}{\method{GloVe}}
\newcommand{\dls}{\method{DLS}}
\newcommand{\sgshort}{\method{sg}}
\newcommand{\sg}{\method{skip-gram}}
\newcommand{\cbow}{\method{cbow}}
\newcommand{\dataset}[1]{\textsc{#1}\xspace}
\newcommand{\apnews}{\dataset{ap-news}}
\newcommand{\wiki}{\dataset{wiki}}
\newcommand{\gnews}{\dataset{gl-news}}
\newcommand{\ppdb}{\dataset{ppdb}}
\newcommand{\bookcorpus}{\dataset{book-corpus}}

\newcommand{\qdup}{{Q-Dup}\xspace}
\newcommand{\textsim}{{STS}\xspace}

\newcommand{\smallurl}[1]{{\smaller{\url{#1}}}}

\newcommand{\secref}[2][]{Section#1~\ref{sec:#2}}

\newcommand{\tabref}[2][]{Table#1~\ref{tab:#2}}
\newcommand{\figref}[2][]{Figure#1~\ref{fig:#2}}

\newcommand{\term}[1]{``#1''\xspace}
\newcommand{\forum}[1]{{\textsl{#1}}\xspace}
\newcommand{\domain}[1]{{\textsl{#1}}\xspace}
\newcommand{\ex}[1]{\textit{#1}\xspace}

\newcommand\email{\begingroup \urlstyle{tt}\smaller[2]\Url}

\aclfinalcopy 

\title{An Empirical Evaluation of \doctovec with \\Practical Insights 
  into
  Document Embedding Generation}

\author{Jey Han Lau$^{1,2}$ \and Timothy Baldwin$^{2}$ \\
    $^1$ IBM Research \\
    $^2$ Dept of Computing and Information Systems,\\The University of 
Melbourne \\
    \email{jeyhan.lau@gmail.com}, \email{tb@ldwin.net}}

\date{}

\begin{document}

\maketitle

\begin{abstract}
  Recently, \newcite{Le+:2014} proposed \doctovec as an extension to
  \wordtovec \cite{Mikolov+:2013b} to learn document-level embeddings.
  Despite promising results in the original paper, others have struggled
  to reproduce those results.  This paper presents a rigorous empirical
  evaluation of \doctovec over two tasks.  We compare \doctovec to two
  baselines and two state-of-the-art document embedding
  methodologies. We found that \doctovec performs robustly when using
  models trained on large external corpora, and can be further improved
  by using pre-trained word embeddings. We also provide recommendations
  on hyper-parameter settings for general-purpose applications, and
  release source code to induce document embeddings using our trained
  \doctovec models.
\end{abstract}

\section{Introduction}

Neural embeddings were first proposed by \newcite{Bengio+:2003}, in the
form of a feed-forward neural network language model.  Modern 
methods use a simpler and more efficient neural architecture to learn
word vectors  (\wordtovec: \newcite{Mikolov+:2013c}; \glove: 
\newcite{Pennington+:2014}), based on objective functions that are designed 
specifically to produce high-quality vectors. 

Neural embeddings learnt by these methods have been applied in a myriad
of NLP applications, including initialising neural network models for 
objective visual recognition \cite{Frome+:2013}
or machine translation \cite{Zhang+:2014a,Li+:2014a},
as well as directly modelling word-to-word relationships 
\cite{Mikolov+:2013b,Zhao+:2015,Salehi+:2015a,Vylomova+:2016a-pre}, 

Paragraph vectors, or \doctovec, were proposed by \newcite{Le+:2014} as
a simple extension to \wordtovec to extend the learning of embeddings
from words to word sequences.\footnote{The term \doctovec was
  popularised by Gensim \cite{gensim}, a widely-used implementation of 
  paragraph
  vectors: \smallurl{https://radimrehurek.com/gensim/}} \doctovec is 
agnostic
to the granularity of the word sequence --- it can equally be a word
$n$-gram, sentence, paragraph or document. In this paper, we use the
term \term{document embedding} to refer to the embedding of a word
sequence, irrespective of its granularity.

\doctovec was proposed in two forms: \dbow and \dmpv. \dbow is a simpler
model and ignores word order, while \dmpv is a more complex model with
more parameters (see \secref{related-work} for details). Although
\newcite{Le+:2014} found that as a standalone method \dmpv is a better
model, others have reported contradictory results.\footnote{The authors of
  Gensim found \dbow outperforms \dmpv:
  \smallurl{https://github.com/piskvorky/gensim/blob/develop/docs/notebooks/doc2vec-IMDB.ipynb}}
\doctovec has also been reported to produce sub-par performance compared
to vector averaging methods based on informal
experiments.\footnote{\smallurl{https://groups.google.com/forum/#!topic/gensim/bEskaT45fXQ}}
Additionally, while \newcite{Le+:2014} report state-of-the-art results
over a sentiment analysis task using \doctovec, others (including the
second author of the original paper in follow-up work) have struggled to replicate this
result.\footnote{For a detailed discussion on replicating the results of
  \newcite{Le+:2014}, see:
  \smallurl{https://groups.google.com/forum/#!topic/word2vec-toolkit/Q49FIrNOQRo}}

Given this background of uncertainty regarding the true effectiveness of
\doctovec and confusion about performance differences between \dbow and 
\dmpv, we aim to shed light on a number of empirical questions: (1) how
effective is \doctovec in different task settings?; (2) which is better
out of \dmpv and \dbow?; (3) is it possible to improve \doctovec through careful 
hyper-parameter optimisation or with pre-trained word embeddings?; and 
(4) can \doctovec be used as an off-the-shelf model like \wordtovec? To 
this end, we present a formal and rigorous evaluation of \doctovec over 
two extrinsic tasks. Our findings reveal that \dbow, despite being the  
simpler model, is superior to \dmpv.  When trained over large external 
corpora, with pre-trained word embeddings and hyper-parameter tuning, we 
find that \doctovec performs very strongly compared to both a simple 
word embedding averaging and $n$-gram baseline, as well as two 
state-of-the-art document embedding approaches, and that \doctovec 
performs particularly strongly over longer documents.  We additionally 
release source code for replicating our experiments, and for inducing 
document embeddings using our trained models.



\section{Related Work}
\label{sec:related-work}

\wordtovec was proposed as an efficient neural approach to learning
high-quality embeddings for words \cite{Mikolov+:2013b}.  Negative
sampling was subsequently introduced as an alternative to the more 
complex hierarchical softmax step at the output layer, with the authors 
finding that not only is it more efficient, but actually produces better 
word vectors on average \cite{Mikolov+:2013c}.

The objective function of \wordtovec is to maximise the log probability 
of context word ($w_O$) given its input word ($w_I$), i.e.\  $\log 
P(w_O|w_I)$. With negative sampling, the objective is to maximise the 
dot product of the $w_I$ and $w_O$ while minimising the dot product of 
$w_I$ and randomly sampled ``negative'' words.  Formally, $\log 
P(w_O|w_I)$ is given as follows:
\begin{multline}
    \log \sigma(v'_{w_O} {}^\intercal v_{w_I}) +  \\
    \sum_{i=1}^k {w_i} \sim P_n(w) \Big[ \log \sigma(-v'_{w_i} 
{}^\intercal v_{w_I}) \Big]
\end{multline}
where $\sigma$ is the sigmoid function, $k$ is the number of negative 
samples, $P_n(w)$ is the noise distribution, $v_{w}$ is the vector of 
word $w$, and $v'_{w}$ is the negative sample vector of word $w$.

There are two approaches within \wordtovec: \sg (``\sgshort'') and 
\cbow. In \sg, the input is a word (i.e.\ $v_{w_I}$ is a vector of 
\textit{one} word) and the output is a context word.  For each input 
word, the number of left or right context words to predict is defined by 
the window size hyper-parameter.  \cbow is different to \sg in one 
aspect: the input consists of multiple words that are combined via 
vector addition to predict the context word (i.e.\  $v_{w_I}$ is a 
summed vector of \textit{several} words).

\doctovec is an extension to \wordtovec for learning document embeddings
\cite{Le+:2014}.  There are two approaches within \doctovec: \dbow and 
\dmpv.

\dbow works in the same way as \sg, except that the input is replaced by
a special token representing the document (i.e.\ $v_{w_I}$ is a vector 
representing the document). In this architecture, the
order of words in the document is ignored; hence the name
\textit{distributed bag of words}.

\dmpv works in a similar way to \cbow. For the input, \dmpv introduces
an additional document token in addition to multiple target words.
Unlike \cbow, however, these vectors are not summed but
concatenated (i.e.\ $v_{w_I}$ is a concatenated vector containing the 
document token and several target words). The objective is again to 
predict a context word given the
concatenated document and word vectors..

More recently, \newcite{Kiros+:2015} proposed \skipthought as a means of
learning document embeddings.  \skipthought vectors are inspired by
abstracting the distributional hypothesis from the word level to the
sentence level.  Using an encoder-decoder neural network architecture,
the encoder learns a dense vector presentation of a sentence, and the
decoder takes this encoding and decodes it by predicting words of its
next (or previous) sentence.  Both the encoder and decoder use a gated
recurrent neural network language model. Evaluating over a range of
tasks, the authors found that \skipthought vectors perform very well
against state-of-the-art task-optimised methods.

\newcite{Wieting+:2016-pre} proposed a more direct way of learning
document embeddings, based on a large-scale training set of paraphrase
pairs from the Paraphrase Database (\ppdb:
\newcite{Ganitkevitch+:2013}).  Given a paraphrase pair, word embeddings 
and a method to compose the word embeddings for a sentence embedding, 
the objective function of the neural network model is to optimise the 
word embeddings such that the cosine similarity of the sentence 
embeddings for the pair is maximised. The authors explore several 
methods of combining word embeddings, and found that simple averaging 
produces the best performance.

\section{Evaluation Tasks}
\label{sec:eval-tasks}

We evaluate \doctovec in two task settings, specifically chosen to
highlight the impact of document length on model performance.

For all tasks, we split the dataset into 2 partitions: development and
test.  The development set is used to optimise the hyper-parameters of
\doctovec, and results are reported on the test set.  We use all 
documents in the development and test set (and potentially more 
background documents, where explicitly mentioned) to train \doctovec.  
Our rationale for this is that the \doctovec training is
completely unsupervised, i.e.\ the model takes only raw text and uses no
supervised or annotated information, and thus there is no need to hold
out the test data, as it is unlabelled. We ultimately relax this 
assumption in the next section (\secref{training-large}), when we train 
\doctovec using large external corpora.

After training \doctovec, document embeddings are generated by the 
model.  For the \wordtovec baseline, we compute a document embedding by 
taking the component-wise mean of its component word embeddings.  We 
experiment with both variants of \doctovec (\dbow and \dmpv) and 
\wordtovec (\sg and \cbow) for all tasks.

In addition to \wordtovec, we experiment with another baseline model 
that converts a document into a distribution over words via maximum 
likelihood estimation, and compute pairwise document similarity using 
the Jensen Shannon divergence.\footnote{We multiply the divergence value 
by $-1.0$ to invert the value, so that a higher value indicates greater 
similarity.} For word types we explore $n$-grams of order $n = 
\{1,2,3,4\}$ and find that a combination of unigrams, bigrams and trigrams
achieves the best results.\footnote{That is, the probability distribution is 
computed over the union of unigrams, bigrams and trigrams in the paired 
documents.} Henceforth, this second baseline will be referred to as \ngram.

\subsection{Forum Question Duplication}
\label{sec:q-dup}

We first evaluate \doctovec over the task of duplicate question detection
in a web forum setting, using the dataset of \newcite{Hoogeveen+:2015}.
The dataset has 12 subforums extracted from StackExchange, and provides
training and test splits in two experimental settings: retrieval and
classification. We use the classification setting, where the goal is to
classify whether a given question pair is a duplicate.

The dataset is separated into the 12 subforums, with a pre-compiled
training--test split per subforum; the total number of instances
(question pairs) ranges from 50M to 1B pairs for the training
partitions, and 30M to 300M pairs for the test partitions, depending on
the subforum.  The proportion of true duplicate pairs is very small in
each subforum, but the setup is intended to respect the distribution
of true duplicate pairs in a real-world setting.

We sub-sample the test partition to create a smaller test partition that
has 10M document pairs.\footnote{Uniform random sampling is used so as
  to respect the original distribution.} On average across all twelve
subforums, there are 22 true positive pairs per 10M question pairs.  We
also create a smaller development partition from the training partition
by randomly selecting 300 positive and 3000 negative pairs. We optimise
the hyper-parameters of \doctovec and \wordtovec using the development
partition on the \forum{tex} subforum, and apply the same hyper-parameter
settings for all subforums when evaluating over the test pairs.  We use
both the question title and body as document content: on average the
test document length is approximately 130 words. We use the default
tokenised and lowercased words given by the dataset. All test,
development and un-sampled documents are pooled together during model
training, and each subforum is trained separately.

We compute cosine similarity between documents using the vectors
produced by \doctovec and \wordtovec to score a document pair.  We then
sort the document pairs in descending order of similarity score, and
evaluate using the area under the curve (AUC) of the receiver operating
characteristic (ROC) curve . The ROC curve tracks the true positive rate
against the false positive rate at each point of the ranking, and as
such works well for heavily-skewed datasets. An AUC score of 1.0 implies
that all true positive pairs are ranked before true negative pairs,
while an AUC score of .5 indicates a random ranking. We present the
full results for each subforum in \tabref{cqa-small}.

Comparing \doctovec and \wordtovec to \ngram, both embedding methods 
perform substantially better in most domains, with two exceptions 
(\forum{english} and \forum{gis}), where \ngram has comparable 
performance.

\doctovec outperforms \wordtovec embeddings in all subforums except for
\forum{gis}. Despite the skewed distribution, simple cosine similarity
based on \doctovec embeddings is able to detect these duplicate
document pairs with a high degree of accuracy.  \dbow performs better than
or as well as \dmpv in 9 out of the 12 subforums, showing that the
simpler \dbow is superior to \dmpv.

One interesting exception is the \forum{english} subforum, where \dmpv
is substantially better, and \ngram~--- which uses only surface word forms 
--- also performs very well.  We hypothesise that the order and the 
surface form of words possibly has a stronger role in this subforum, as 
questions are often about grammar problems and as such the position and 
semantics of words is less predictable (e.g.\ \ex{Where does ``for the 
same'' come from?})

\begin{table}[t]
\begin{center}
\begin{adjustbox}{max width=0.45\textwidth}
\begin{tabular}{c|c@{\,\,}c:c@{\,\,}c:c}
\multirow{2}{*}{\textbf{Subforum}} & 
\multicolumn{2}{c:}{\textbf{\doctovec}} & 
\multicolumn{2}{c:}{\textbf{\wordtovec}} & 
\multirow{2}{*}{\textbf{\ngram}} \\
& \dbow & \dmpv & \sgshort & \cbow & \\
\hline
\forum{android} & \textbf{.97} & .96 & .86 & .93 & .80 \\
\forum{english} & .84 & \textbf{.90}  & .76 & .73 & .84 \\
\forum{gaming} & \textbf{1.00}  & .98 & .97 & .97 & .94\\
\forum{gis} & .93 & .95 & .94 & \textbf{.97} & .92 \\
\forum{mathematica} & \textbf{.96}  & .90 & .81 & .81 & .70\\
\forum{physics} & .96 & \textbf{.99}  & .93 & .90 & .88\\
\forum{programmers} & \textbf{.93}  & .83 & .84 & .84 & .68\\
\forum{stats} & \textbf{1.00}  & .95 & .91 & .88 & .77\\
\forum{tex} & \textbf{.94}  & .91 & .79 & .86 & .78\\
\forum{unix} & \textbf{.98}  & .95 & .91 & .91 & .75\\
\forum{webmasters} & \textbf{.92}  & .91 & \textbf{.92}  & .90 & .79\\
\forum{wordpress} & \textbf{.97}  & \textbf{.97}  & .79 & .84 & .87\\
\end{tabular}
\end{adjustbox}
\end{center}
\caption{ROC AUC scores for each subforum. Boldface indicates the best 
 score in each row.}
\label{tab:cqa-small}
\end{table}

\subsection{Semantic Textual Similarity}
\label{sec:sts}

\begin{table}[t!]
\begin{center}
\begin{adjustbox}{max width=0.47\textwidth}
\begin{tabular}{c|c:c@{\,\,}c:c@{\,\,}c:c}
\multirow{2}{*}{\textbf{Domain}} & \multirow{2}{*}{\textbf{\dls}} & 
\multicolumn{2}{c:}{\textbf{\doctovec}} & 
\multicolumn{2}{c:}{\textbf{\wordtovec}} & 
\multirow{2}{*}{\textbf{\ngram}} \\
& & \dbow & \dmpv & \sgshort & \cbow & \\
\hline
\domain{headlines} & .83 & .77 & \textbf{.78} & .74 & .69 & .61 \\
\domain{ans-forums} & .74 & \textbf{.66} & .65 & .62 & .52 & .50 \\
\domain{ans-students} & .77 & .65 & .60 & \textbf{.69} & .64 & .65 \\
\domain{belief} & .74 & \textbf{.76} & .75 & .72 & .59 & .67 \\
\domain{images} & .86 & \textbf{.78} & .75 & .73 & .69 & .62\\
\end{tabular}
\end{adjustbox}
\end{center}
\caption{Pearson's $r$ of the STS task across 5 domains. \dls is the 
overall best system in the competition. Boldface indicates the best 
results between \doctovec and \wordtovec in each row.}
\label{tab:sts-small}
\end{table}

\begin{table}[t]
\footnotesize
\begin{center}
\begin{tabular}{rp{4cm}}
\textbf{Hyper-Parameter} & \textbf{Description} \\
\hline
Vector Size & Dimension of word vectors \\
Window Size & Left/right context window size \\
Min Count & Minimum frequency threshold for word types \\
Sub-sampling & Threshold to downsample high frequency words \\
Negative Sample & No. of negative word samples \\
Epoch & Number of training epochs \\
\end{tabular}
\end{center}
\caption{A description of \doctovec hyper-paramters.}
\label{tab:hyperparameter-description}
\end{table}

\begin{table*}[t]
\begin{center}
\begin{adjustbox}{max width=0.95\textwidth}
\begin{tabular}{ccccccccc}
\multirow{2}{*}{\textbf{Method}} & \multirow{2}{*}{\textbf{Task}} & 
\textbf{Training} & \textbf{Vector} & \textbf{Window} & \textbf{Min} & 
\textbf{Sub-} & \textbf{Negative} & \multirow{2}{*}{\textbf{Epoch}} \\
&& \textbf{Size} & \textbf{Size} & \textbf{Size} & \textbf{Count} & 
\textbf{Sampling} & \textbf{Sample} & \\
\hline
\multirow{2}{*}{\dbow} & \qdup & 4.3M & 300 & 15 & 5 & 10$^{-5}$ & 5 & 20\\
& \textsim & .5M & 300 & 15 & 1 & 10$^{-5}$ & 5 & 400\\
\hdashline
\multirow{2}{*}{\dmpv}& \qdup & 4.3M & 300 & 5 & 5 & 10$^{-6}$ & 5 & 600\\
& \textsim & .5M & 300 & 5 & 1 & 10$^{-6}$ & 5 & 1000\\
\end{tabular}
\end{adjustbox}
\end{center}
\caption{Optimal \doctovec hyper-parameter values used for each tasks.  
``Training size'' is the total word count in the training data. For 
\qdup training size is an average word count across all subforums.}
\label{tab:hyperparameter-values}
\end{table*}

The Semantic Textual Similarity (STS) task is a shared task held as part
of *SEM and SemEval over a number of iterations
\cite{Agirre+:2013,Agirre+:2014,Eneko+:2015}. In STS, the goal is to
automatically predict the similarity of a pair of sentences in the range
$[0,5]$, where 0 indicates no similarity whatsoever and 5 indicates
semantic equivalence.

The top systems utilise word alignment, and further 
optimise their scores using supervised learning \cite{Eneko+:2015}. Word 
embeddings are employed, although sentence embeddings are often taken as 
the average of word embeddings (e.g.\ \newcite{Sultan+:2015}).

We evaluate \doctovec and \wordtovec embeddings over the English STS sub-task of 
SemEval-2015 \cite{Eneko+:2015}. The dataset has 5 domains, and each 
domain has 375--750 annotated pairs. Sentences are much shorter than our 
previous task, at an average of only 13 words in each test sentence.

As the dataset is also much smaller, we combine sentences from all 5 
domains and also sentences from previous years (2012--2014) to form the 
training data.  We use the \domain{headlines} domain from 2014 as development, 
and test on all 2015 domains. For pre-processing, we tokenise and 
lowercase the words using Stanford CoreNLP \cite{Manning+:2014}.

As a benchmark, we include results from the overall top-performing
system in the competition, referred to as ``\dls'' \cite{Sultan+:2015}.
Note, however, that this system is supervised and highly customised to
the task, whereas our methods are completely unsupervised. Results are
presented in \tabref{sts-small}.

Unsurprisingly, we do not exceed the overall performance of the
supervised benchmark system \dls, although \doctovec outperforms \dls
over the domain of \domain{belief}. \ngram performs substantially worse 
than all methods (with an exception in \forum{ans-students} where it 
outperforms \dmpv and \cbow).

Comparing \doctovec and \wordtovec, \doctovec performs better.  However, 
the performance gap is lower compared to the previous two tasks, 
suggesting that the benefit of using \doctovec is diminished for shorter 
documents. Comparing \dbow and \dmpv, the difference is marginal, 
although \dbow as a whole is slightly stronger, consistent with the 
observation of previous task.

\subsection{Optimal Hyper-parameter Settings}
\label{sec:optimal-hyperparameter}

Across the two tasks, we found that the optimal hyper-parameter
settings (as described in \tabref{hyperparameter-description}) are
fairly consistent for \dbow and \dmpv, as detailed in
\tabref{hyperparameter-values} (task abbreviations: 
\qdup $=$ Forum Question Duplication (\secref{q-dup}); and \textsim $=$ 
Semantic Textual Similarity (\secref{sts})). Note that we did not tune 
the initial and minimum learning rates ($\alpha$ and $\alpha_{min}$, 
respectively), and use the the following values for all experiments: 
$\alpha = .025$ and $\alpha_{min} = .0001$.  The learning rate decreases 
linearly per epoch from the initial rate to the minimum rate.

In general, \dbow favours longer windows for context words than
\dmpv. Possibly the most important hyper-parameter is the sub-sampling
threshold for high frequency words: in our experiments we find that task
performance dips considerably when a sub-optimal value is used.  \dmpv
also requires more training epochs than \dbow.  As a rule of thumb, for
\dmpv to reach convergence, the number of epochs is one order of
magnitude larger than \dbow. Given that \dmpv has more parameters in the
model, this is perhaps not a surprising finding.

\section{Training with Large External Corpora}
\label{sec:training-large}

\begin{table*}[t]
\begin{center}
\begin{adjustbox}{max width=0.95\textwidth}
\begin{tabular}{ccc|c:c:cc:ccc:c}
\multirow{2}{*}{\textbf{Task}} &
\multirow{2}{*}{\textbf{Metric}} &
\multirow{2}{*}{\textbf{Domain}} &
{\textbf{\pp}} &
{\textbf{\skipthought}} &
\multicolumn{2}{c:}{\textbf{\dbow}} & \multicolumn{3}{c:}{\textbf{\sg}} 
& \multirow{2}{*}{{\textbf{\ngram}}} \\
& && \ppdb & \bookcorpus & \wiki & \apnews & \wiki & \apnews & \gnews & 
 \\
\hline
\multirow{12}{*}{\qdup} & \multirow{12}{*}{AUC} & \forum{android} & .92 
& .57 & \textbf{.96} & .94 & .77 & .76 & .72 & .80 \\
&& \forum{english} & {.82} & .56 & .80 & .81 & .62 & .63 & .61 & 
\textbf{.84} \\
&& \forum{gaming} & \textbf{.96} & .70 & .95 & .93 & .88 & .85 & .83 & 
.94 \\
&& \forum{gis} & {.89} & .58 & .85 & .86 & .79 & .83 & .79 & 
\textbf{.92} \\
&& \forum{mathematica} & .80 & .57 & \textbf{.84} & .80 & .65 & .58 & 
.59 & .70 \\
&& \forum{physics} & \textbf{.97} & .61 & .92 & .94 & .81 & .77 & .74 & 
.88 \\
&& \forum{programmers} & .88 & .69 & \textbf{.93} & .88 & .75 & .72 & 
.64 & .68  \\
&& \forum{stats} & .87 & .60 & .92 & \textbf{.98} & .70 & .72 & .66 & 
.77  \\
&& \forum{tex} & .88 & .65 & \textbf{.89} & .82 & .75 & .64 & .73  & .78 
\\
&& \forum{unix} & .86 & .74 & \textbf{.95} & .94 & .78 & .72 & .66  & 
.75  \\
&& \forum{webmasters} & .89 & .53 & .89 & \textbf{.91} & .77 & .73 & .71 
& .79 \\
&& \forum{wordpress} & .83 & .66 & \textbf{.99} & .98 & .61 & .58 & .58 
&  .87 \\
\hdashline
\multirow{5}{*}{\textsim} & \multirow{5}{*}{$r$} & \domain{headlines} &
\textbf{.77} & .44 & .73 & .75 & .73  & .74  & .66 &  .61 \\
&& \domain{ans-forums} & \textbf{.67} & .35 & .59 & .60 & .46 & .44 & 
.42 & .50 \\
&& \domain{ans-students} & \textbf{.78} & .33 & .65 & .69 & .67 & .69 & 
.65  & .65 \\
&& \domain{belief} & \textbf{.78} & .24 & .58 & .62 & .51 & .51 & .52 & 
.67
\\
&& \domain{images} & \textbf{.83} & .18 & .80 & .78 & .72 & .73 & .69 & 
.62 \\

\end{tabular}
\end{adjustbox}
\end{center}
\caption{Results over all two tasks using models trained with external 
corpora.}
\label{tab:train-external}
\end{table*}

In \secref{eval-tasks}, all tasks were trained using small in-domain
document collections.  \doctovec is designed to scale to large data, and 
we explore the effectiveness of \doctovec by training it on large 
external corpora in this section.

We experiment with two external corpora: (1) \wiki, the full collection of English 
Wikipedia;\footnote{Using the dump dated 2015-12-01, cleaned using
  WikiExtractor: \smallurl{https://github.com/attardi/wikiextractor}} 
and (2) \apnews, a collection of Associated Press English news articles from 
2009 to 2015. We tokenise and lowercase the documents using Stanford 
     CoreNLP \cite{Manning+:2014}, and treat each natural paragraph of 
an article as a document for \doctovec. After pre-processing, we have 
approximately 35M documents and 2B tokens for \wiki, and 25M and .9B 
tokens for \apnews. Seeing that \dbow trains faster and is a better 
model than \dmpv from \secref{eval-tasks}, we experiment with only \dbow 
here.\footnote{We use these hyper-parameter values for \wiki (\apnews): 
vector size $=$ 300 (300), window size $=$ 
15 (15), min count $=$ 20 (10), sub-sampling threshold $=$ 10$^{-5}$ 
   (10$^{-5}$), negative sample $=$ 5, epoch $=$ 20 (30). After removing 
low frequency words, the vocabulary size is approximately 670K for \wiki 
and 300K for \apnews.}

To test if \doctovec can be used as an off-the-shelf model, we take a
pre-trained model and infer an embedding for a new document without
updating the hidden layer word weights.\footnote{That is, test data is 
held out and not including as part of \doctovec training.}  We have 
three hyper-parameters for test inference: initial learning rate 
($\alpha$), minimum learning rate ($\alpha_{min}$), and number of 
inference epochs.  We optimise these parameters using the development 
partitions in each task; in general a small initial $\alpha$ (= .01) 
with low $\alpha_{min}$ (= .0001) and large epoch number (= 500--1000) 
works well.

For \wordtovec, we train \sg on the same  
corpora.\footnote{Hyper-parameter values for \wiki (\apnews): vector 
size $=$ 300 (300), window size $=$ 5 (5), min count $=$ 20 (10), 
sub-sampling threshold $=$ 10$^{-5}$ (10$^{-5}$), negative sample $=$ 5, 
epoch $=$ 100 (150)}  We also include the word vectors trained on the 
larger Google News by \newcite{Mikolov+:2013c}, which has 100B 
words.\footnote{\smallurl{https://code.google.com/archive/p/word2vec/}}  
The Google News \sg vectors will henceforth be referred to as \gnews.

\dbow, \sg and \ngram results for all two tasks are presented in 
\tabref{train-external}. Between the baselines \ngram and \sg, \ngram 
appears to do better over \qdup, while \sg works better over \textsim.

As before, \doctovec outperforms \wordtovec and \ngram across almost all 
tasks.  For tasks with longer documents (\qdup), the performance gap 
between \doctovec and \wordtovec is more pronounced, while for \textsim, 
which has shorter documents, the gap is smaller. In some \textsim 
domains (e.g.\ \domain{ans-students}) \wordtovec performs just as well 
as \doctovec.  Interestingly, we see that \gnews \xspace  \wordtovec 
embeddings perform worse than our \wiki and \apnews  \xspace \wordtovec 
embeddings, even though the Google News corpus is orders of magnitude 
larger.

Comparing \doctovec results with in-domain results 
(\ref{tab:cqa-small} and \ref{tab:sts-small}), the performance is in 
general lower. As a whole, the performance difference between the \dbow 
models trained using \wiki and \apnews is not very large, indicating the 
robustness of these large external corpora for general-purpose 
applications. To facilitate applications using off-the-shelf \doctovec 
models, we have publicly released code and trained models to induce 
document embeddings using the \wiki and \apnews \dbow  
models.\footnote{\smallurl{https://github.com/jhlau/doc2vec}}


\subsection{Comparison with Other Document Embedding Methodologies}

We next calibrate the results for \doctovec against \skipthought
\cite{Kiros+:2015} and paragram-phrase (\pp: 
\newcite{Wieting+:2016-pre}), two recently-proposed competitor document 
embedding methods. For \skipthought, we use the pre-trained model made 
available by the authors, based on the \bookcorpus dataset 
\cite{Zhu+:2015}; for \pp, once again we use the pre-trained model from 
the authors, based on \ppdb \cite{Ganitkevitch+:2013}. We compare these 
two models against \dbow trained on each of \wiki and \apnews.  The 
results are presented in \tabref{train-external}, along with results for 
the baseline method of \sg and \ngram.

\skipthought performs poorly: its performance is worse than the simpler
method of \wordtovec vector averaging and \ngram. \dbow outperforms \pp 
over most \qdup subforums, although the situation is reversed for 
\textsim.  Given that \pp is based on word vector averaging, these
observations support the conclusion that vector averaging methods works best for
shorter documents, while \dbow handles longer documents better.

It is worth noting that \doctovec has the upperhand compared to \pp in 
that it can be trained on in-domain documents. If we compare in-domain 
\doctovec results 
(\ref{tab:cqa-small} and \ref{tab:sts-small}) to \pp 
(\tabref{train-external}), the performance gain on \qdup is even more 
pronounced.


\section{Improving \doctovec with Pre-trained Word Embeddings}
\label{sec:improve-pretrain}

Although not explicitly mentioned in the original paper \cite{Le+:2014}, 
\dbow does not learn embeddings for words in the default configuration.
In its implementation (e.g.\ Gensim), 
\dbow  has an option to turn on word embedding learning, by running a 
step of \sg to update word embeddings before running \dbow.
With the option turned off, word embeddings are randomly initialised and 
kept at these randomised values.

Even though \dbow can in theory work with randomised word embeddings, we
found that performance degrades severely under this setting. An
intuitive explanation can be traced back to its objective function,
which is to maximise the dot product between the document embedding and
its constituent word embeddings: if word embeddings are randomly
distributed, it becomes more difficult to optimise the document
embedding to be close to its more critical content words.

\begin{figure*}[t]
        \centering
        \begin{subfigure}[b]{.5\textwidth}
                \includegraphics[width=\textwidth]{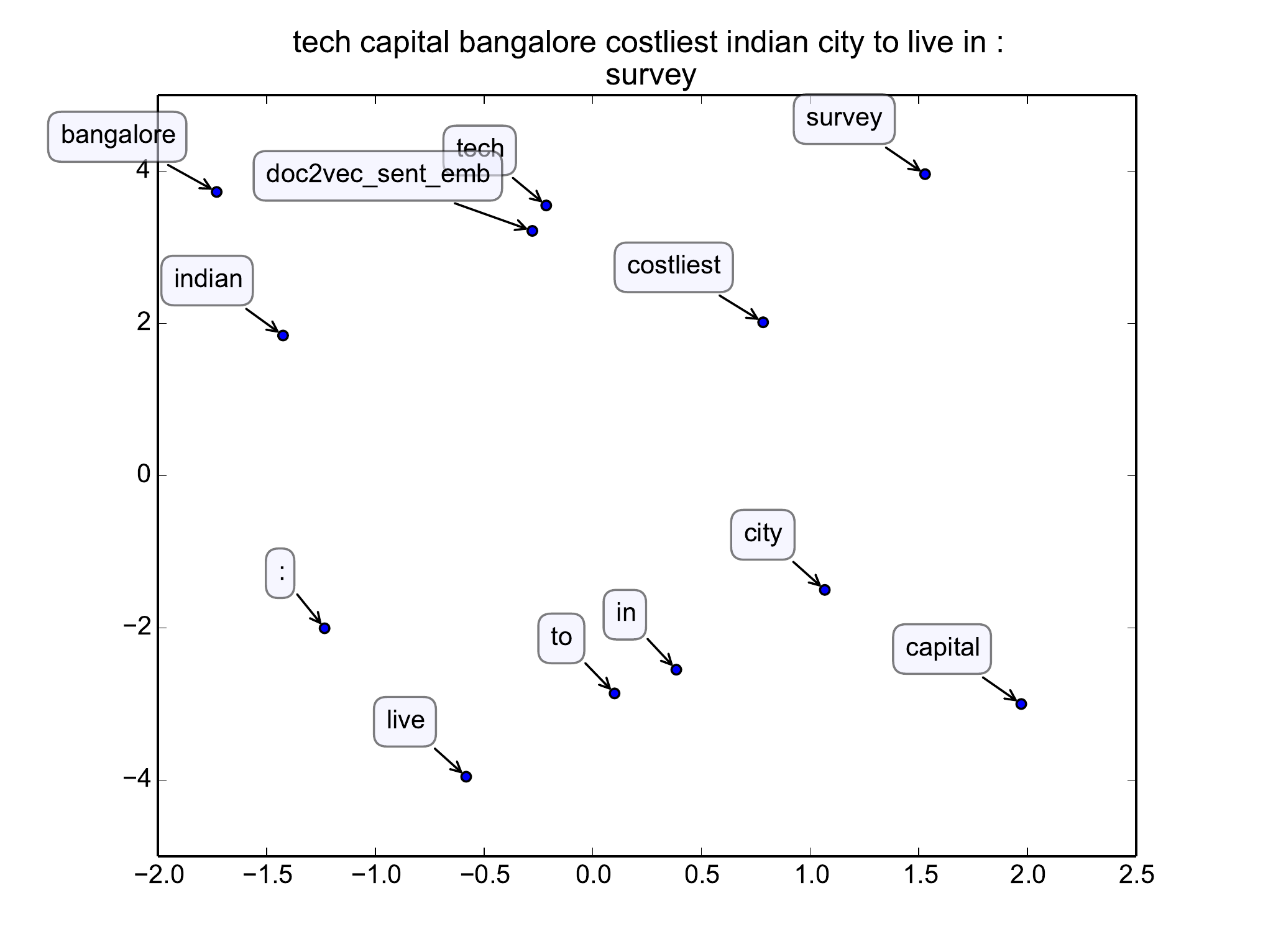}
                \caption{\doctovec (\dbow)}
        \end{subfigure}%
        ~
        \begin{subfigure}[b]{.5\textwidth}
                \includegraphics[width=\textwidth]{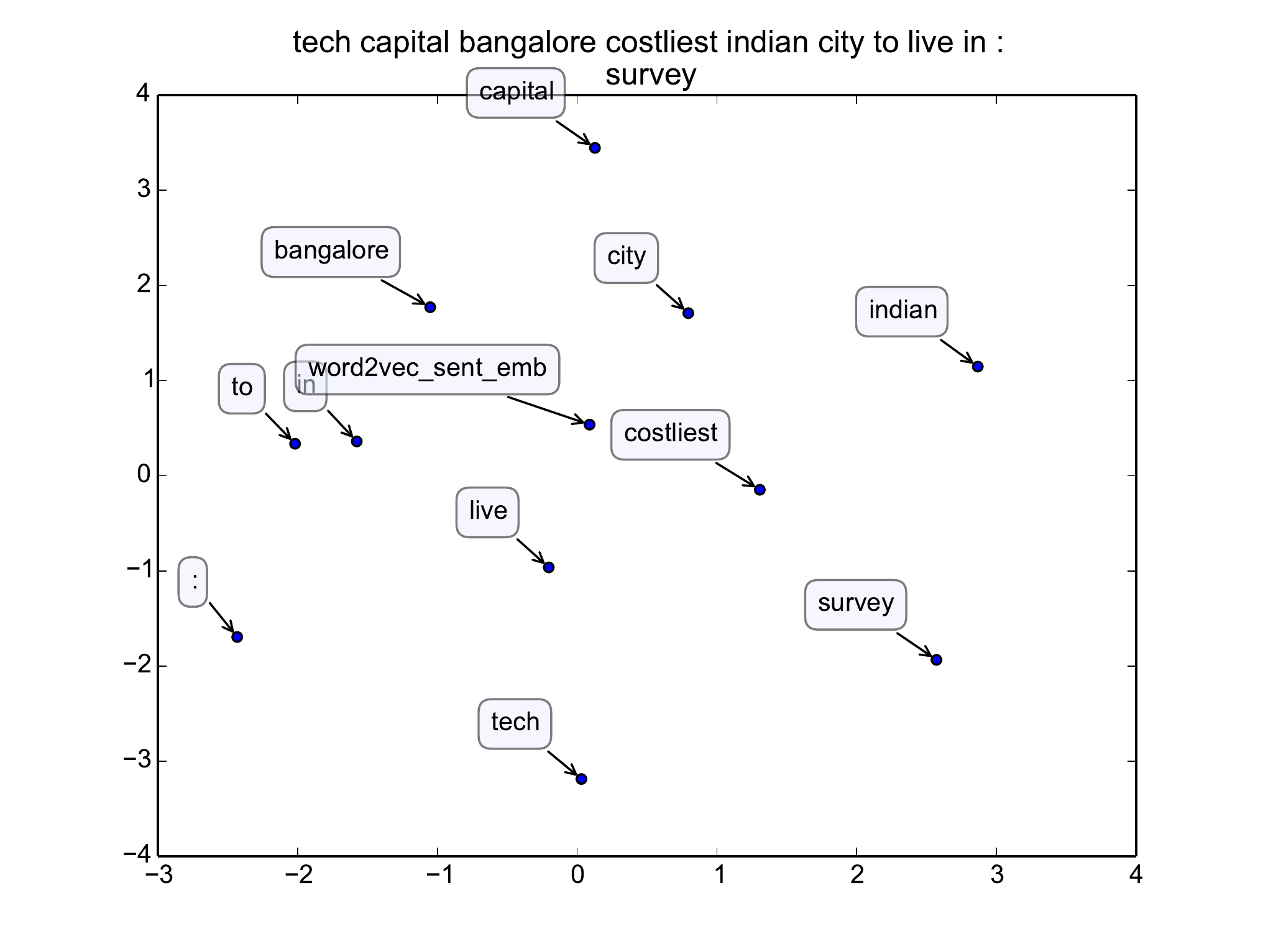}
                \caption{\wordtovec (\sg)}
        \end{subfigure}
        \caption{Two-dimentional t-SNE projection of \doctovec and 
\wordtovec embeddings.}
\label{fig:2d-plot}
\end{figure*}

To illustrate this, consider the two-dimensional t-SNE plot
\cite{vanderMaaten:Hinto:2008} of \doctovec document and word embeddings
in \figref{2d-plot}(a).  In this case, the word learning option is
turned on, and related words form clusters, allowing the document
embedding to selectively position itself closer to a particular word
cluster (e.g.\ content words) and distance itself from other
clusters (e.g.\ function words).  If word embeddings are randomly
distributed on the plane, it would be harder to optimise the document
embedding.

\begin{table}
\begin{center}
\begin{adjustbox}{max width=0.45\textwidth}
\begin{tabular}{cc|ccc}
\multirow{2}{*}{\textbf{Task}} &
\multirow{2}{*}{\textbf{Domain}} &
\multirow{2}{*}{\textbf{\dbow}} & \textbf{\dbow$+$} & \textbf{\dbow$+$} 
\\
&&&\wiki & \apnews \\
\hline
\multirow{12}{*}{\qdup} & 
\forum{android} &
.97 & \textbf{.99} & .98   \\
& \forum{english} & .84 & \textbf{.90} & .89   \\
& \forum{gaming} & \textbf{1.00} & \textbf{1.00} & \textbf{1.00}   \\
& \forum{gis} & .93 & .92 & \textbf{.94}   \\
& \forum{mathematica} & \textbf{.96} & \textbf{.96} & \textbf{.96}  \\
& \forum{physics} & .96 & \textbf{.98} & .97   \\
& \forum{programmers} & \textbf{.93} & .92 & .91  \\
& \forum{stats} & \textbf{1.00} & \textbf{1.00} & .99   \\
& \forum{tex} & .94 & \textbf{.95} & .92   \\
& \forum{unix} & \textbf{.98} & \textbf{.98} & .97  \\
& \forum{webmasters} & .92 & \textbf{.93} & \textbf{.93}   \\
& \forum{wordpress} & .97 & .96 & \textbf{.98}   \\
\hdashline
\multirow{5}{*}{\textsim} & 
\domain{headlines} &
.77 & \textbf{.78}& \textbf{.78}    \\
& \domain{ans-forums} & .66 & \textbf{.68} & \textbf{.68}   \\
& \domain{ans-students} & \textbf{.65} & .63 & \textbf{.65}   \\
& \domain{belief} & .76 & .77 & \textbf{.78}  \\
& \domain{images} & .78 & \textbf{.80} & .79   \\
\end{tabular}
\end{adjustbox}
\end{center}
\caption{Comparison of \dbow performance using pre-trained \wiki and 
\apnews \xspace \sg embeddings.}
\label{tab:pretrain}
\end{table}

Seeing that word vectors are essentially learnt via \sg in \dbow, we 
explore the possibility of using externally trained \sg word embeddings 
to initialise the word embeddings in \dbow. We repeat the experiments 
described in \secref{eval-tasks}, training the \dbow model using the 
smaller in-domain document collections in each task, but this time 
initialise the word vectors using pre-trained \wordtovec embeddings from 
\wiki and \apnews.  The motivation is that with better initialisation, 
the model could converge faster and improve the quality of embeddings.

Results using pre-trained \wiki and \apnews \xspace \sg embeddings are 
presented in \tabref{pretrain}. Encouragingly, we see that using 
pre-trained word embeddings helps the training of \dbow on the smaller 
in-domain document collection.  Across all tasks, we see an increase in 
performance.  More importantly, using pre-trained word embeddings 
\textit{never} harms the performance.  
Although not detailed in the table, we also find that the
number of epochs to achieve optimal performance (based on development data) 
is fewer than before.

We also experimented with using pre-trained \cbow word embeddings for 
\dbow, and found similar observations. This suggests that the 
initialisation of word embeddings of \dbow is not sensitive to a 
particular word embedding  implementation.

\section{Discussion}
\label{sec:discussion}

To date, we have focused on quantitative evaluation of \doctovec and
\wordtovec. The qualitative difference between \doctovec and \wordtovec
document embeddings, however, remains unclear. To shed light on what is
being learned, we select a random document from \textsim~--- \ex{tech
  capital bangalore costliest indian city to live in: survey} --- and
plot the document and word embeddings induced by \dbow and
\sg using t-SNE in \figref{2d-plot}.\footnote{We plotted a larger
  set of sentences as part of this analysis, and found that the general
  trend was the same across all sentences.}

For \wordtovec, the document embedding is a centroid of the word
embeddings, given the simple word averaging method. With \doctovec, on
the other hand, the document embedding is clearly biased towards the
content words such as \ex{tech}, \ex{costliest} and \ex{bangalore}, and
away from the function words.  \doctovec learns this from its objective
function with negative sampling: high frequency function words are 
likely to be selected as negative samples, and so the document embedding 
will tend to align itself with lower frequency content words.


\section{Conclusion}
\label{sec:conclusion}

We used two tasks to empirically evaluate the quality of document
embeddings learnt by \doctovec, as compared to two baseline methods --- 
\wordtovec word vector averaging and an $n$-gram model --- and two 
competitor document embedding methodologies.  Overall, we found that 
\doctovec performs well, and that \dbow is a better model than \dmpv. We 
empirically arrived at recommendations on optimal \doctovec 
hyper-parameter settings for general-purpose applications, and found 
that \doctovec performs robustly even when trained using large external 
corpora, and benefits from pre-trained word embeddings. To facilitate 
the use of \doctovec and enable replication of these results, we release 
our code and pre-trained models.

\bibliographystyle{acl2016}
\bibliography{strings,citations,papers}

\end{document}